\documentclass{article}

\usepackage[final]{neurips_2020_tda}

\usepackage[utf8]{inputenc} 
\usepackage[T1]{fontenc}    
\usepackage{amsfonts}       
\usepackage{booktabs}       
\usepackage{hyperref}       
\usepackage{url}            
\usepackage{microtype}      
\usepackage{nicefrac}       
\usepackage{paralist}       
\usepackage{siunitx}        
\usepackage{wrapfig}
\usepackage[utf8]{inputenc} 
\usepackage[T1]{fontenc}    
\usepackage{hyperref}       
\usepackage{url}            
\usepackage{booktabs}       
\usepackage{amsfonts}       
\usepackage{nicefrac}       
\usepackage{microtype}      

\usepackage{graphicx}
\usepackage{color,soul}
\usepackage{amsmath}
\usepackage{amsthm}
\usepackage{amstext}
\usepackage{mathrsfs}
\usepackage{amssymb}
\usepackage{bm}
\usepackage{pictexwd,dcpic}

\usepackage{tikz}
\usetikzlibrary{calc,intersections,through}
\tikzset{
    partial ellipse/.style args={#1:#2:#3}{
        insert path={+ (#1:#3) arc (#1:#2:#3)}
    }
}

\newtheorem{theorem}{Theorem}[section]

\theoremstyle{definition}

\newtheorem{proposition}[theorem]{Proposition}
\newtheorem{example}[theorem]{Example}

\newtheorem{remark}[theorem]{Remark}

\newcommand{\D}{\mathcal{D}} 
\newcommand{\M}{\mathcal{M}} 
\newcommand{\R}{\mathbb{R}} 
\newcommand{\Z}{\mathbb{Z}} 
\newcommand{\e}{\epsilon} 
\newcommand{\A}{\mathcal{A}} 

\title{Learning a manifold from a teacher's demonstrations}

 \author{
   Pei Wang\\
Rutgers University--Newark\\
\texttt{peiwang@rutgers.edu} \\
 \And
Arash Givchi \\
Rutgers University--Newark\\
\texttt{arash.givchi@gmail.com} \\
\And
Patrick Shafto\\
Rutgers University--Newark\\
\texttt{patrick.shafto@gmail.com} \\
}

\begin{document}

\maketitle

\begin{abstract}
 We consider the problem of learning a manifold from a teacher's demonstration. Extending existing approaches of learning from randomly 
 sampled data points, we consider contexts where data may be chosen by a teacher. 
 We analyze learning from teachers who can provide structured data such as individual examples (isolated data points) and demonstrations (sequences of points). Our analysis shows that for the purpose of teaching the topology of a manifold, demonstrations can yield 
 remarkable decreases in the amount of data points required in comparison to teaching with randomly sampled points.
We also discuss the implications of our analysis for learning in humans and machines. 
\end{abstract}

\section{Introduction}

In machine learning, learners are assumed to operate in a relatively simplified problem space: data points sampled by a random process. Humans learn from a richer, stronger context. 
Two aspects that have received the most attention are the fact that data may be chosen by a more knowledgeable informant, such as a teacher \cite{Shafto2014,Zhu2015}, and that the data points may themselves be structured into sequences as in demonstration \cite{kuhl1997cross,brand2002evidence}. In this paper, we consider how having teachers who select structured data may affect the complexity of learning.

Our goal is to understand theoretical bounds on learning manifolds and their topology from teaching via structured data, which we expect to inform debates in machine learning and human learning. 
We particularly investigate the topology of manifolds for several reasons.  
First, the decomposition of learning into grounded and more abstract aspects parallels common wisdom across human and machine learning, which have converged on hierarchical (``deep'') models of learning. Second, teaching topology will prove to be data-efficient for manifold learning applications, such as clustering where only information about the global structure of manifold (e.g.  number of connected components etc.)  is needed.  Third, teaching will be able to proceed without full knowledge of the geometry, and the requirement for the teacher can be relaxed by just knowing the homotopy type of the manifold.

We begin with preliminaries in Sec.~\ref{sec:prelim}. Sec.~\ref{sec:teachtopology} provides results related to teaching the topology of manifolds via data points and demonstrations, showing that demonstrations can yield vastly more efficient teaching. 
Sec.~\ref{sec:conclusion} provides concluding discussions.


\section{Preliminaries}\label{sec:prelim}


In machine learning, the manifold assumption states that high dimensional data in the real world are typically concentrated on a much lower dimensional manifold.
Because the difficulty of inferring the geometry of an arbitrary manifold is bounded by its worst local feature, quantified as the \textit{reach} of the manifold, we may only aim to reduce the sample complexity of learning a manifold by focusing on its global properties, which are encoded by the topology. 

In this paper, $\M$ is an orientable compact sub-manifold in $\R^n$.
We mainly focus on low dimensional manifolds such as curves and surfaces.
However, teaching methods developed in the following sections can be directly used to
convey low dimensional topological features of any manifold. 

For the formalism of teaching-learning algorithms, we consider $\mathcal A$ as a class of \textbf{learning algorithms} that construct approximations of $\M$ and/or identify the homotopy type of $\M$ from a set of data points sampled from $\M$. Examples of such algorithms are available in \cite{cheng2005manifold, niyogi2008finding,boissonnat2014manifold}.

A collection of data points $ \D \subset \M$ is called a \textbf{teaching set} with respect to $\mathcal A$ if there exists a learning algorithm $A \in \mathcal A$ that recovers the homotopy type of $\M$ using $\D$. $|\D|$ denotes the size of $\D$. A teaching set $\D^*$ is said to be \textit{minimal w.r.t. $\M$} if $|\D^*|\leq |\D|$ for any teaching set $\D$ of $\M$.
Further, $\D^*$ is a \textit{minimal teaching set w.r.t. the homotopy type of $\M$} if $\D^*$ is a teaching set for some $\M'$ of the same homotopy type as $\M$ and $|\D^*|\leq |\D|$ for any teaching set $\D$ of a manifold homotopy equivalent to $\M$. 
The size of a minimal teaching set is called the \textbf{minimal teaching number}. 

\section{Structured data and manifold teaching}\label{sec:teachtopology}

\subsection{Manifold teaching from sample points} \label{sec:points}
The pioneering work in \cite{niyogi2008finding} introduced a framework to reconstruct manifolds from random sampling.
Their work can be rephrased as a manifold teaching problem. Suppose two agents, which we call a teacher and a learner, wish to communicate a manifold $\M \subset \R^n$. In their setting, the teacher passes a collection of {\em randomly} sampled data points $\D=\{x_1, \dots, x_k\} $ to the learner, who then builds a manifold by a learning algorithm in the \textbf{class $\A(\e)$}:  the learner first picks a parameter~$\e \in \R^{+}$, then for each $x_i \in \D$, makes an n-dimensional ball $B_{\e}(x_i)$ centered at $x_i$ of radius~$\e$. Here $n$ is the dimension of the ambient space which can be inferred from data points' coordinate size. The union of all these balls $U_{\e}(\D)=\cup_{x\in \D}B_{\e}(x)$ constitutes the learned space. 

The main result in \cite{niyogi2008finding} provides an estimation $N_{\e}$ on the number of data that are needed to guarantee that the learned space $U_{\e}$ and the target manifold $\M$ are homotopy equivalent with a given confidence. $N_{\e}$ depends on the confidence level, the volume and the reach of $\M$, and also the learner's choice of $\e$. 

Considering $N_{\e}$ as a sufficient bound on the minimal teaching number of $\M$, we seek a necessary condition. The calculation of $N_{\e}$ proposed in \cite{niyogi2008finding} requires knowledge about the geometric features of $\M$ (volume and reach), which translated to our context implies that either the teacher knows the true $\M$ or the teacher has observed a large amount of data points, which allows one to make good estimations of these geometric features. 
Hence we will start our analysis by assuming that the teacher has access to the true manifold. 
In Section~\ref{sec:conclusion}, we will discuss how this assumption can be relaxed in many practical cases.

\begin{wrapfigure}[10]{r}{0.2\textwidth}
 \centering
(a)\includegraphics[width=1in, height=0.75in]{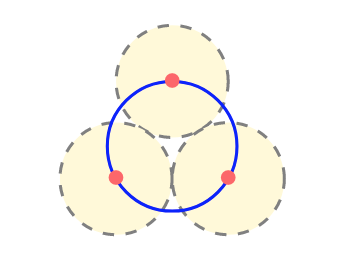}
(b)\includegraphics[width=1in, height=0.75in]{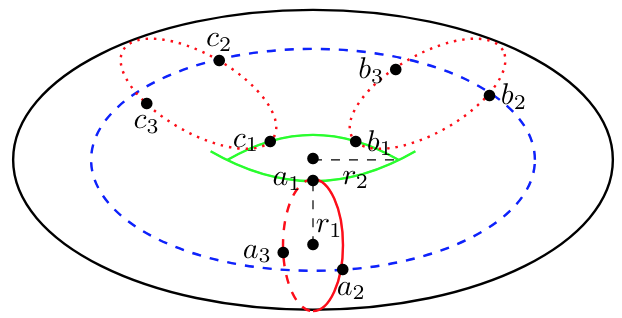}
\end{wrapfigure}

Suppose that the learner uses the class of algorithms $\A(\e)$, what is a \textit{minimal teaching set} to convey the homotopy type of $\M$? 
The case for non-contractible 1-dim manifolds are extremely neat. Since every such $\M$ is homotopy equivalent to a circle, at least three points are needed as explained~below.

\begin{example} \label{eg:circle}
Let $S^{1}$ be a unit circle embedded in $\mathbb R^2$, and $\mathcal A(\epsilon)$ be the class of learning algorithms described above. 
It is clear that any data set with only one or two points will result contractible $U_{\e}$ for any choice of~$\epsilon$.
However, as illustrated in Figure~(a), with three equidistant points sampled from $S^1$, any learner $\A(\e)$ with $\frac{\sqrt{3}}{2}\leq \e < 1$ will recover the correct topology of $S^1$ from the union of three connected disks with a hole in the middle. Thus the minimal teaching number for a circle is three. 
\end{example}

Suppose that $\M$ is a closed orientable surface. An example is given below.

\begin{example} \label{eg: torus}
Let $T^2$ be a torus embedded in $\R^3$ as shown in Figure~(b).
$T^2$ can be obtained by rotating the \textcolor{red}{red} circle $l_{1}$ around the \textcolor{green}{green} circle $l_{2}$.
Denote the radii of $l_{1}$ and $l_{2}$ by $r_{1}$ and $r_{2}$ respectively.
Two 1-dim holes of $T^2$ are represented by $l_1$ and $l_{2}$.
As in Example~\ref{eg:circle}, each $l_i$ needs at least three teaching points. 
Since the learner $\A(\e)$ picks one $\e$ for all data points, more data points are needed for $l_i$ when $\frac{\sqrt{3}}{2}r_i\geq r_j$, where
$i, j\in \{1, 2\}$ and $i\neq j$. Hence to find the minimal teaching set for the homotopy type of $T^2$, we may assume that $r_1=r_2=r$.
Suppose that any three data points sharing a circle in Figure~(b) are equidistant points.
Then $\D_1=\{a_1, a_2, a_3, b_1, c_1\}$ can be used to teach $l_{1}$ and $l_2$. 
To recover the only 2-dim hole of $T^2$, it is natural to add $\{b_2, b_3, c_2, c_3\}$ into $\D_1$ to complete the red dotted circles going through $b_1$ and $c_1$. Ideally, $\e$-balls centered at these $9$ points should form a torus. However, there are large undesirable gaps left open between the red circles because the learner is restricted to pick $\frac{\sqrt{3}}{2} r<\e<r$.

We now compute how many extra points are needed to fill in all these gaps. Direct calculation shows that the radius of the dashed blue circle $l_3$ is $2.5 r$ and nine equidistant data points on $l_3$ are needed to teach it with $\frac{\sqrt{3}}{2} r<\e<r$. 
If we rotate $l_1$ around $l_2$ nine times with each step $2\pi/9$, then the trace of $\{a_1, a_2, a_3\}$ produces $27$ data points (including all 9 points in $\D_1$). With these $27$ points, we almost form a torus but still have many small gaps. 
One may count that in total there are $27$ such gaps. So $54$ points are enough. Moreover, notice that the inner green circle $l_2$ is over taught, one may check that 3 teaching points can be removed from $l_2$. Hence we may teach $T^2$ with $51$ points. 
\end{example}

The approach we used in Example~\ref{eg: torus} can be generalized to all orientable surfaces. 

\begin{proposition}\label{prop:closedsurface}
Let $\M_g \subset \R^3$ be a closed orientable surface with genus $g$. Then the \textit{minimal teaching number} for the \textit{homotopy type} of $\M_g$ 
with respect to $\A(\e)$ is bounded by $49g+2$.
\end{proposition}
\textit{Proof.} We will proceed by induction. When $g=1$, $\M_1$ is homotopy equivalent to $T^2$. So the homotopy type of $\M_1$ can be taught by 51 points. 
Suppose that the claim holds for any $\M_g$ with $g<n$. 
Then when $g<n$, $M_{g,1}$, surface with genus $g$ and $1$ boundary component, can be taught by $49g+1$ points.
Notice that there exists a $\M_n$ which can be obtained by gluing a $M_{1,1}$ with a $\M_{n-1,1}$. Hence we may teach $\M_n$ with $[49 + 1] + [49(n-1)+1] = 49n+2$ data points. $\Box$


\begin{remark}
Let $\M_{g,b} \subset \R^3$ be a genus $g$ orientable surface with $b$ boundary components. Note that $\M_{g, b}$ can be obtained from $\M_{g}$ by removing $b$ disconnect disks. Therefore, the \textit{minimal teaching number} of $\M_{g,b}$ with respect to $\A(\e)$ is bounded by $49g+2-b$.
\end{remark}

The above analysis suggests that due to the locally Euclidean nature of manifolds, 
a local to global teaching procedure as described above is not always efficient: even a simple manifold as a regular torus requires a large set of teaching points. Below we propose a new class of teaching algorithms that teaches the topology directly from demonstrations where each demonstration is a sequence of data describing a loop.




\subsection{Manifold teaching from demonstrations}\label{sec:seq}


\begin{wrapfigure}[19]{r}{0.2\textwidth}
  \begin{center}
    (c)\includegraphics[width=1.1in, height=0.75in]{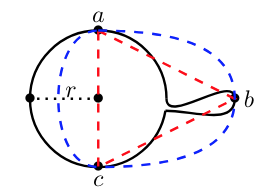}
    (d)\includegraphics[width=1.1in, height=0.8in]{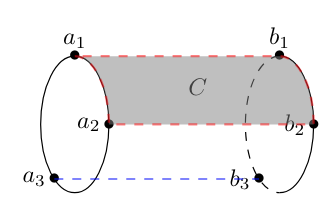}
    (e)\includegraphics[width=1.1in, height=0.8in]{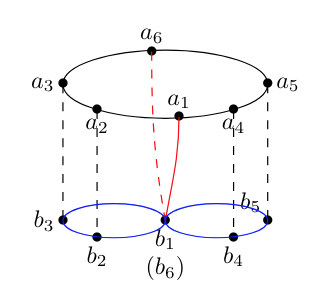}
  \end{center}
\end{wrapfigure}

Teaching with sequential data is efficient because topologies of manifolds are intuitively captured by loops in various dimensions. 
As in Example~\ref{eg:circle}, a unit circle can be taught by three points. In fact, any (oriented) \footnote{Without considering the orientation, a sequence with two points can also describe a loop.} non-contractible 1-dim manifold can be effectively described by a sequence consisting three \textit{randomly sampled} points. For example, the teacher may teach the black loop $\M_b$ in Figure~(c) by $\D=\{[a, b, c, a]\}$. The sequential data informs the learner to connect consecutive points by a simple curve\footnote{A curve is simple if it has no self-intersection.}, which will form a space $U_{\D}$ that is homotopy equivalent to $\M_b$. The red and blue curves in Figure~(c) are two examples of $U_{\D}$ obtained by different learners.
If curves that connect data points further adopt some mild assumption, for example smoothness, polynomial, linear etc., then the obtained space $U_{\D}$ can be parameterized accordingly. 

    

    


\begin{remark}
It is important to note that calculations in this section are done for a particular target manifold $\M$, whereas in the previous section were done for the homotopy type of $\M$.
For instance, given an arbitrary non-contractible 1-dim manifold $\M^1$, any sequence with three points randomly sampled from $\M^1$ forms a teaching set for $\M^1$. However for learners using $\A(\e)$ (Sec~\ref{sec:points}), it is true that the minimal teaching size for the homotopy type of $\M^1$ w.r.t. $\A(\e)$ is three, but it is possible that there does not exist three points on $\M^1$ form a teaching set for $\M^1$ based on $\A(\e)$. 
\end{remark}





Higher dimensional manifolds can also be conveyed using sequential data.
In this setting, the teacher passes a \textit{sequence of sequences} to the learner. The learner builds a manifold by a learning algorithm in the \textbf{class $\A(\mathbf{l})$} where $\mathbf{l}$ indexes all the choices: for each sequence $\mathbf{a}=[a_1, \dots, a_m]$, the learner connects consecutive points by a curve; for each sequence of sequences $[\mathbf{a}, \mathbf{b}, \dots, \mathbf{c}]$ where $\mathbf{a}, \mathbf{b}, \dots, \mathbf{c}$ are sequences of data points with the same length, the learner connects consecutive sequences by curved planes (the curvature is not necessarily zero). In more details, as Figure~(d), with $\mathbf{a}=[a_1, a_2, a_3, a_1], \mathbf{b}=[b_1, b_2, b_3, b_1]$, the learner first connects points in $\mathbf{a}$ and $\mathbf{b}$ separately to form two loops (shown as black circles). Furthermore, to join $\mathbf{a}$ and $\mathbf{b}$, the learner may link each distinct pair of $a_i$ and $b_i$ by a curve \footnote{If $a_i$ and $b_i$ are the same point, do nothing.} (shown as dished lines). This completes a closed path through $a_i, a_{i+1}, b_{i+1}$\footnote{$a_{4}=a_{1}$ and $b_4=b_1$.} and $ b_i$, for $i=1,2,3$. Then the learner glues a curved plane along each of these closed path. For example, the \textcolor{gray}{gray} area $C$ in Figure~(d) shows a curved plane glued along the \textcolor{red}{red} closed path going through $a_1, a_2, b_2, b_1$.

To make the learning algorithm $\A(\mathbf{l})$ robust to the choice of curves and planes, we further assume that the teacher and the learner agree that $(*)$: (1) there is no intersection between different connecting curves except end points; (2) two points are connected by at most one curve. 
For instance, a pair of pants can be taught by 
$\D_{\text{pants}}={ \{ \scriptstyle{[[a_1, a_2, a_3, a_6][b_1, b_2, b_3, b_6]], [[a_1, a_4, a_5, a_6][b_1, b_4, b_5, b_6]]}\}}$ as shown in Figure~(e). With the teaching set $\D_{\text{pants}}$, the learner needs to connect $a_1, b_1$ and $a_6, b_6$ multiple times. If the learner makes the connection by the \textcolor{red}{red} curves for the first time, assumption~$(*)$ ensures the learner always picks the red curves during the entire learning process.

\begin{example}\label{eg:toruspair}
The torus in Example~\ref{eg: torus} can be taught by a sequence of four sequences: 

$\D_{\text{torus}}=[[a_1, a_2, a_3, a_1] [b_1, b_2, b_3, b_1] [c_1, c_2, c_3, c_1] [a_1, a_2, a_3, a_1]]$ as shown in Figure~(b). 
This teaching set only contains the 9 basic points which fits our initial intuition. 
\end{example}


\begin{proposition}
Let $\M_g \subset \R^3$ be a closed orientable surface with genus $g\geq 2$. Then the \textit{minimal teaching number} of $\M_g$ 
with respect to $\A(\mathbf{l})$ is bounded by $3g-3$ sequences, where each sequence consists of at most $4$ data points.
\end{proposition}
\textit{Proof.} A classical result of surfaces states that for any $\M_g$, there is a system of $3g-3$ disjoint simple closed curves which cut $\M_g$ into pairs of pants (see for example, \cite{farb2011primer}). Note that each simple closed curve can be taught by a sequence of $4$ points; each pair of pants can be taught by a sequence that consists of three sequences representing its boundary curves. 
Moreover, two legs of a pair of pants can be glued along their boundary curves through a sequential data. For instance, two \textcolor{blue}{blue} boundary curves in Figure~(e) can be glued by $[[b_1, b_2, b_3, b_1][b_1,b_4, b_5, b_1 ]]$. Hence the claim holds. $\Box$

\section{Conclusions}\label{sec:conclusion}
We considered the problem of teaching low-dimensional features of a manifold using structured data, which extends mathematical approaches to learning manifolds toward the contexts more consistent with the richness of human learning. Building on prior work in manifold learning, we formalize teaching manifolds from data points, observe that contrary to intuition, teaching does not facilitate learning as much as one would expect due to constraints imposed by the reach of the manifold. Considering learning from demonstrations---sequences of data points---we show that learning can be greatly facilitated by teaching. This approach relies on separating teaching the geometry of the manifold itself from teaching its topology. Focusing on teaching only the topology, we show that sequences of points can be used to represent the homology groups of the manifold, which compactly capture important abstract structure that can be used to facilitate future learning. Moreover, this relaxes the overly stringent and implausible requirement that the teacher must know the manifold exactly to an estimation of its homotopy type, which is almost always less stringent than the true manifold.
A preliminary example where the teacher teaches with partial knowledge and unconstrained data is included in the Supplementary material (See Section~\ref{sec: HL}).
Finally, we argue for a connection to human learning from teaching demonstrations, which are most naturally thought of as sequences. 
Future work may extend this approach toward more naturalistic learning problems faced by humans or solved by machine learning. The approaches are not restricted to manifold teaching and it would be interesting to explore teaching more general mathematical objects with low dimensional topological structures, such as graphs, CW-complexes and even groups.

\newpage





\bibliographystyle{plainnat} 
\bibliography{references}

\newpage

\appendix

 \section{Learning from teacher with partial knowledge}\label{sec: HL}


All teaching methods discussed in the main text assume teacher has full access to the true manifold. 
However, in reality, the teacher often does not know the underlying manifold and often does not have full control over which data can be used to teach.
In this section, we consider teaching in a much more practical scenario that allows a teacher, who may have limited knowledge, to teach with unconstrained data. We illustrate how this would assist the learner to improve their estimation of the relevant topological and geometrical information from the data.

Following the standard setting of \textit{topological data analysis} \cite{carlsson2009topology, chazal2017introduction}, we assume that the data $\D$ is a finite set of points sampled from 
the true manifold $\M$. 
Using an algorithm in \textbf{class $\A(\e)$} (Sec~\ref{sec:points}) with different $\e$'s, the learner obtains a summary of estimations of $\M$ in form of \textit{persistent homology}. 
Rather than picking a teaching set directly from $\M$,
the teacher first selects a subset $\D_{T}$ from $\D$, then passes $\D_{T}$ to the learner in a proper sequential format according to algorithm $\A(\mathbf{l})$ (Sec~\ref{sec:seq}) to demonstrate desired topological features of $\M$.

As discussed before, the difficulty of learning a manifold $\M$ increases dramatically as the reach of $\M$ drops.
Now we illustrate how teaching helps in these situations by the following example. 

\begin{figure}[h]
    \centering
    \includegraphics[scale=0.3]{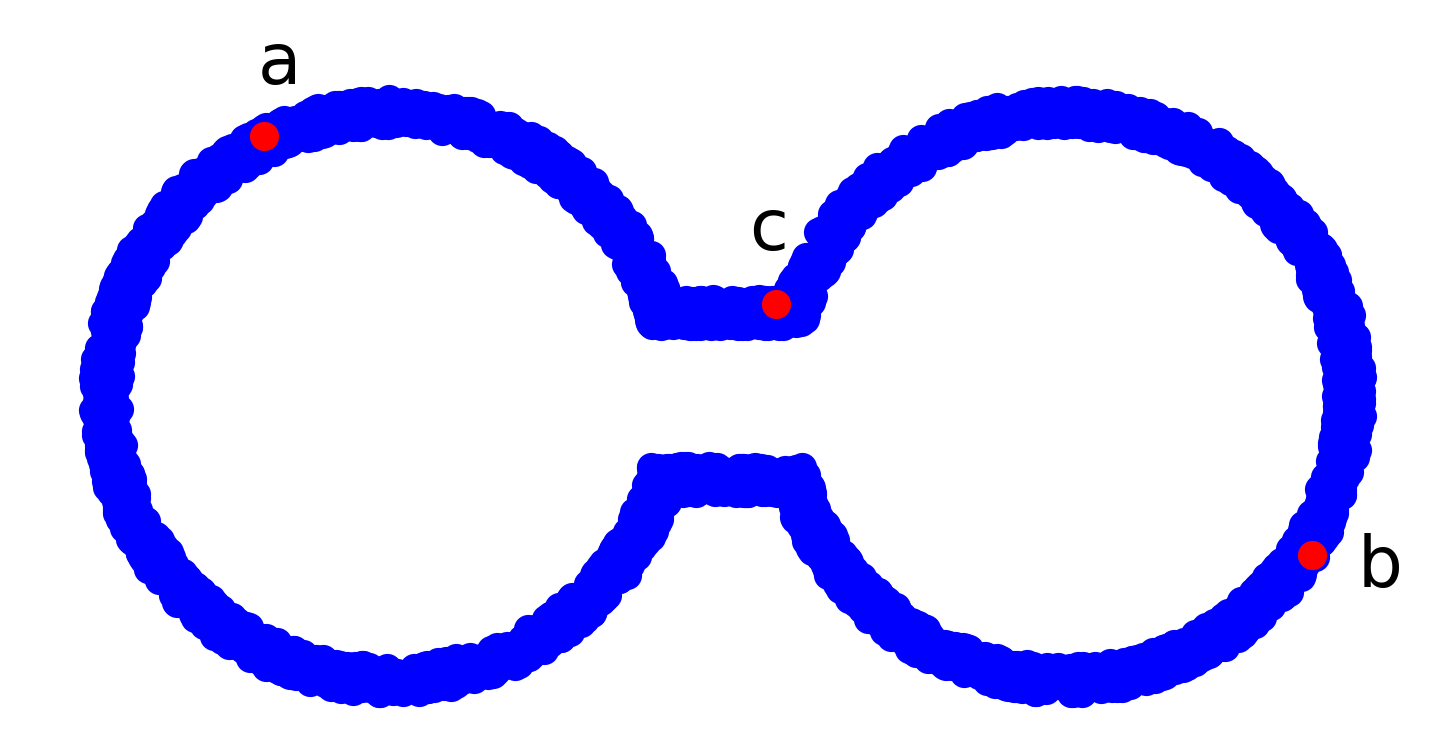}
    \caption{A barbell shaped annulus.}
    \label{fig:barbell}
\end{figure}

\begin{figure}[ht]
    \centering
 
   (i)\includegraphics[scale=0.4]{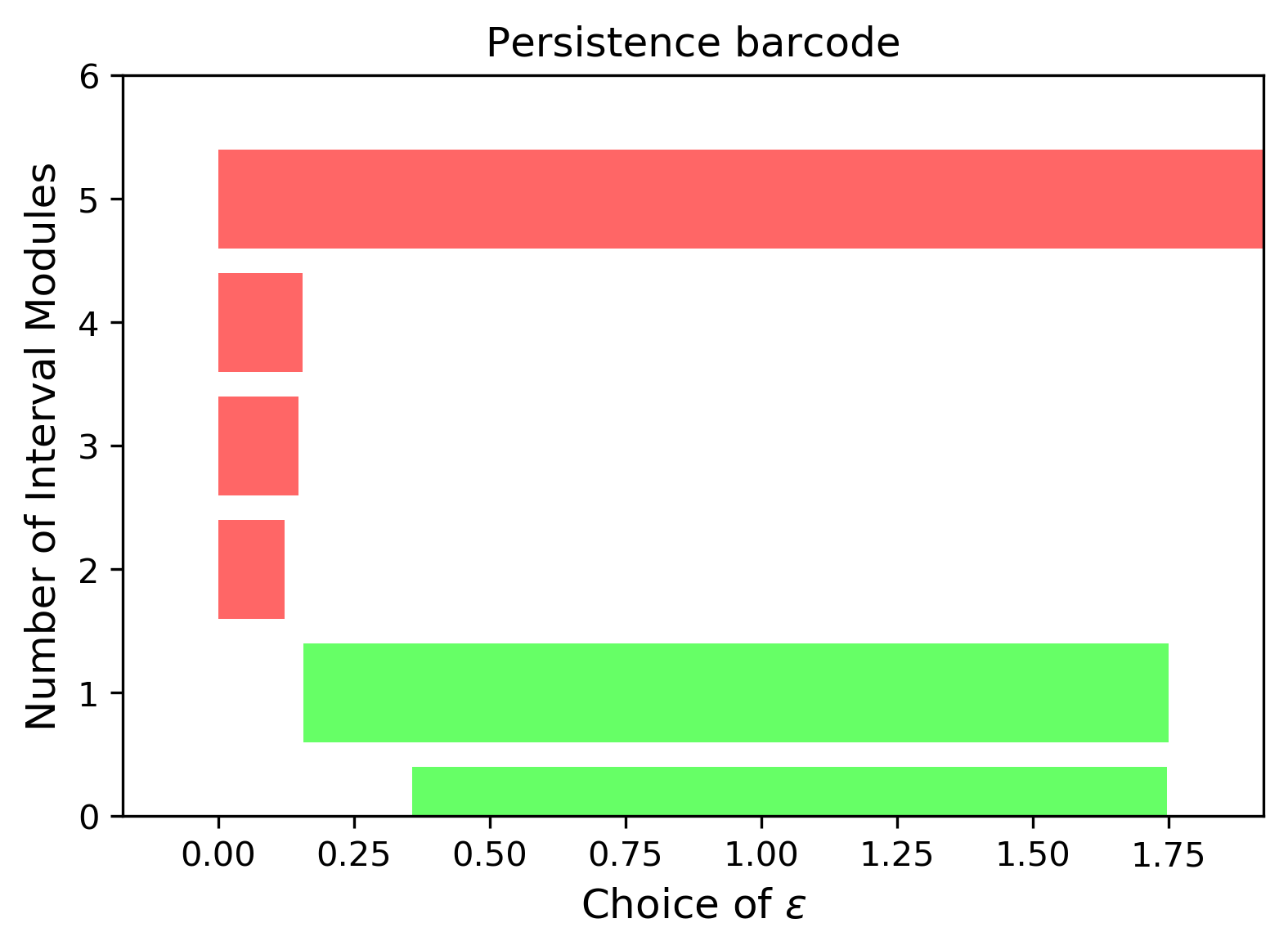}
   \hspace{0.2in}
   (ii)\includegraphics[scale=0.4]{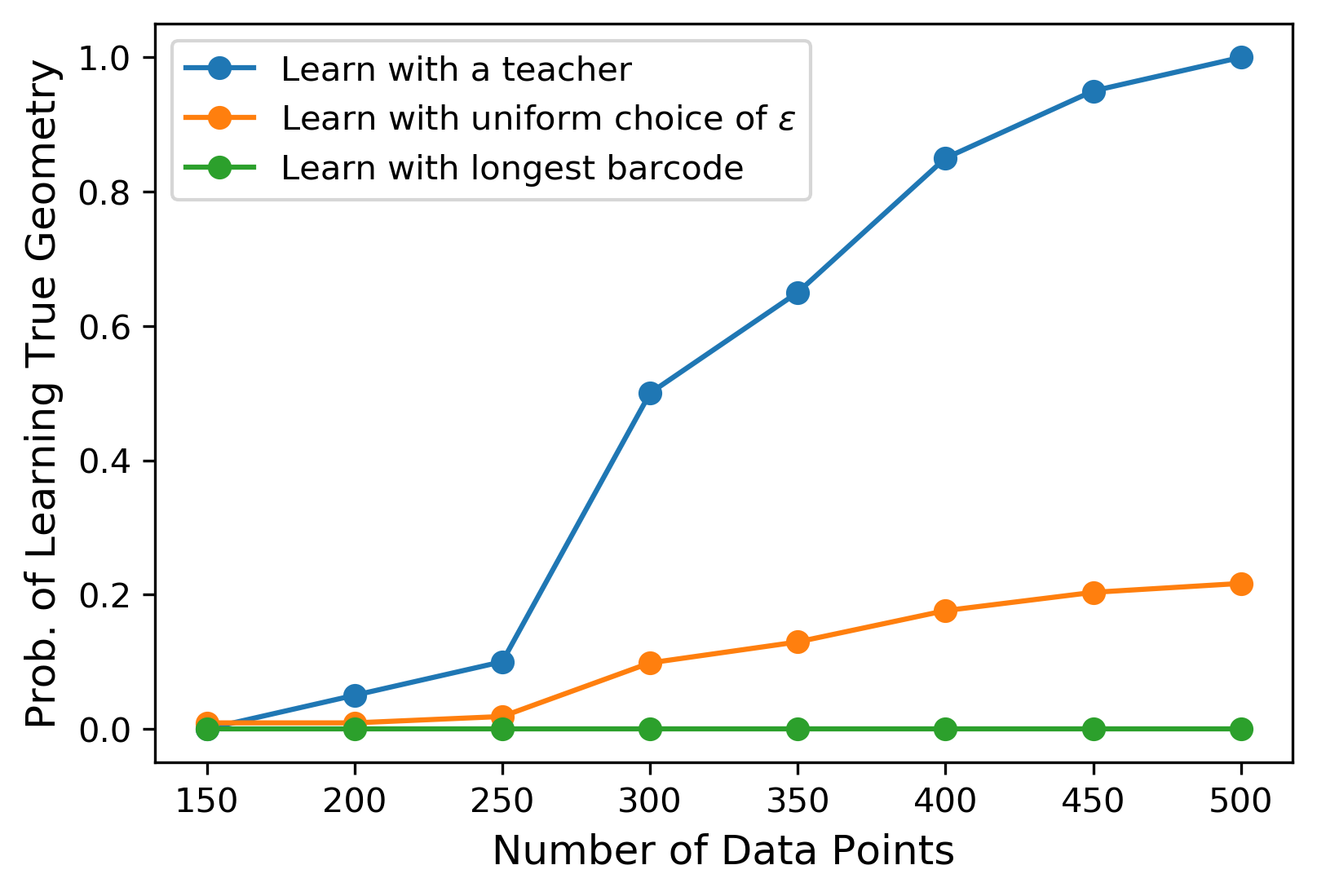}
    
    \caption{Learning based on TDA } \label{fig:dense-barbell}
\end{figure}

\begin{example}\label{eg:limited_teaching}
Let the true manifold $\M$ be the \textcolor{blue}{blue} barbell shaped annulus shown in
Figure~\ref{fig:barbell} with reach $\tau = 0.26$.
Assume that the learner analyzes randomly sampled data by TDA and 
the teacher knows that $\M$ contains a 1-dim hole.
Based on $\A(\mathbf{l})$, three distinct points are required to form a teaching sequence for this hole.
When fewer than three data points are observed by the learner, the teacher would simply wait until more data were collected.
Suppose that the learner gets three data $\D_{1}=\{a, b, c\}$ as shown.
The corresponding persistence barcode of $\D_{1}$  is empty for $H_{1}$ (no 1-dim loop is ever formed for any choice of $\e$).
With $\D_{1}$,  the teacher may teach by marking these points sequentially as for example $[a, b, c, a]$.
Comparing the teacher's demonstration with the barcode, 
the learner would realize that $\M$ is homotopy equivalent to a circle and currently points gathered 
are not sufficient to extract any accurate geometrical information.

Further suppose that the learner intends to estimate the geometry of $\M$ and so more points are sampled. 
A given data set $\D$ is called \textit{feasible}, 
if the learner is able to derive the true geometry of $\M$ from $\D$ with some~$\e$, i.e.
if there exists $\e^*<\tau = 0.26$ such that $U_{\e*}(\D)$ is homotopy equivalent to $\M$
\footnote{It is possible that $U_{\e}(\D)$ is homotopy equivalent to $\M$ for $\e>\tau$.
However the top and the bottom of the narrow middle part of $\M$ will be connected up in such $U_{\e}(\D)$, which leads to wrong geometry.}.
To estimate the lower bound on size of a feasible data set,
we randomly sample data sets from $\M$ with increasing sizes and 20 simulations for each size.
Empirically it shows that feasible data sets appear only after $|\D|>150$ and 
and appears in every simulation for $|\D|\geq 500$.

Figure~\ref{fig:dense-barbell}(i) shows 
the persistence barcode for a data set of size $500$.\footnote{The barcode was constructed using the GUDHI library \cite{maria2014gudhi}.}
The \textcolor{red}{red} bars are the longest four intervals for $H_0$, which reflects the number of connected components.
After $\e>0.156$, only one red bar remains which indicates $U_{\e}$ contains a single component for any $\e>0.156$.
The \textcolor{green}{green} bars are the intervals for $H_1$ (ignoring intervals of length less than 0.05), 
which represents the number of 1-dim holes.
The top green bar spans over $(0.158, 1.751)$ and indicates that there is a 1-dim loop forms at $\e=0.158$ and persists until $\e= 1.751$.
The bottom green bar spans over $(0.357, 1.747)$ and indicates that another 1-dim loop forms at $\e=0.357$ and persists until $\e= 1.747$.
All randomly sampled data sets of size $500$ exhibit similar persistence barcode with two long intervals for $H_1$ as shown.
Focusing on the range of $\e$ where $1$-dim holes exist, on average 78\% choice of $\e$ (with variance 0.0002) indicates two $1$-dim loops over all simulations.
Thus, without teaching, the learner would likely to conclude a wrong topological information, $H_1(\M) = \Z_2\times \Z_2$, with high confidence. 
In contrast, with a teaching set of three points, 
the learner is able to not only infer the correct topology immediately after teaching but also 
accurately estimate the geometry of $\M$ by focusing on $U_{\e}$ with $0.158<\e<0.357$.

Figure~\ref{fig:dense-barbell}(ii) plots the average learning accuracy of $\M$'s geometry for different types of learners.
The blue curve shows the learners with a topological teacher who are assumed to follow a Bernoulli distribution since they are able to infer the correct geometry with every feasible data set.
The orange curve is corresponding to learners who choose $\e$ uniformly from the interval where barcode for $H_1(\M)$ is not empty (Variances are omitted as their magnitudes are bounded above by $0.01$).
The green curve shows learners, who approximate $\M$ by $U_{\e}$ with the most persistent homology, stay incorrect on geometry even with increasing data size. 
Clearly learner's acquisition of geometry are accelerated by teaching topology. 
\end{example}

Persistent homology has started to attract attention in machine learning \cite{carlsson2008local, chazal2013persistence,li2014persistence, reininghaus2015stable}.
However, levering these topological features for learning poses considerable challenges because the relevant topological information
is not carried by the whole persistence barcode but is concentrated in a small region of $\e$ that may not be obvious \cite{hofer2017deep}.
Teaching by demonstration resolves these challenges by allowing the the learner to extract the most suitable topological information after the correct homology appears in the persistence barcode, and zooming the analysis of $\M$'s geometry into
the most appropriate range of $\e$ with high data efficiency.





More importantly, teaching by demonstration allows \textit{accumulation} of information across learners, whereas other forms of teaching can only transmit information from an already knowledgeable teacher. As pointed out in Sec~\ref{sec:points}, the method of teaching by sampling points essentially assumes that the teacher knows the true manifold $\M$. However, given the intractability of manifold learning in general, there is no plausible way for the teacher to have access to $\M$. 
On such accounts, teaching does not resolve the true challenge of learning and instead passes off the problem to a teacher for whom the learning problem does not exist. 
The key advantage of teaching from demonstrations is that it allows the teacher to convey critical information of $\M$ without knowing the entire manifold. For example, let $\M$ be a torus as in Figure~(b) with $r_1<<r_2$. The teacher may only have enough observations to conclude that there is a loop homotopy equivalent to the \textcolor{green}{green} circle $l_2$. With sequential data, the teacher could easily pass the only loop $l_2$ he observed, which allows the learner to focus on the region of $\e$ where $l_2$ exists.

In addition, from a teacher's perspective, much less data is needed to learn the topology of an irregular manifold $\M$ than its geometry. 
For instance, let $\M$ be the 1-dim manifold shown in Figure~(c). Denote the reach of $\M$ by $\tau$ and the radius of the left arc in $\M$ by $r$. Note that the teacher only needs $r$-dense data to learn the topology of $\M$, whereas $\tau$-dense data to learn the geometry. In fact, for any manifold $\M$, we may define its \textbf{topological reach} $\eta$ to be the largest number such that $U_{\e}(\M)$ is homotopy equivalent to $\M$ for any $\e\leq \eta$, where $U_{\e}(\M)=\cup_{\{p\in \M\}}B_{\e}(p)$. According to Proposition~3.2 in \cite{niyogi2008finding}, for the same confidence level, points needed to achieve $\e$-dense is polynomial increasing with $1/\e$. Therefore when $\M$ is irregular, i.e. $\tau$ is significantly less than $\eta$, the amount of data needed to achieve $\eta$-dense is much fewer than $\tau$-dense. Since the topology of $U_{\e}(\M)$ remains the same for data beyond $\eta$-dense, 
it requires much less data to learn the topology of an irregular manifold than its geometry. 
\end{document}